\pdfoutput=1

\documentclass[11pt]{article}

\usepackage[preprint]{acl}

\usepackage{times}
\usepackage{latexsym}

\usepackage[T1]{fontenc}

\usepackage[utf8]{inputenc}

\usepackage{microtype}

\usepackage{inconsolata}

\usepackage{graphicx}
\usepackage{multirow}
\usepackage{diagbox}
\usepackage{graphicx}
\usepackage{subcaption}
\usepackage{enumerate}
\usepackage{enumitem} 
\usepackage{rotating}
\usepackage{mathdots}
\usepackage{caption}
\usepackage{wrapfig}
\usepackage{amsmath}
\usepackage{amssymb}
\usepackage{booktabs} 
\title{FoldGPT: Simple and Effective Large Language Model Compression Scheme}

\author{Songwei Liu$^*$~~~
        Chao Zeng\thanks{Equal contribution}~~~
        Lianqiang Li~~~
        Chenqian Yan~~~ \\
        \textbf{Lean Fu}~~~ 
        \textbf{Xing Mei}~~~
        \textbf{Fangmin Chen} \thanks{Corresponding author}\\
        ByteDance Inc. \\
        \texttt{\{ zengchaocs, cfangmin\}@gmail.com,} \\
        \texttt{\{liusongwei.zju, lilianqiang, yanchenqian.i, fulean, xing.mei\}@bytedance.com}}

\begin{document}
\maketitle
\begin{abstract}
The demand for deploying large language models(LLMs) on mobile devices continues to increase, driven by escalating data security concerns and cloud costs. However, network bandwidth and memory limitations pose challenges for deploying billion-level models on mobile devices. In this study, we investigate the outputs of different layers across various scales of LLMs and found that the outputs of most layers exhibit significant similarity. Moreover, this similarity becomes more pronounced as the model size increases, indicating substantial redundancy in the depth direction of the LLMs. Based on this observation, we propose an efficient model volume compression strategy, termed \textbf{FoldGPT}, which combines block removal and block parameter sharing.This strategy consists of three parts: (1) Based on the learnable gating parameters, we determine the block importance ranking while modeling the coupling effect between blocks. Then we delete some redundant layers based on the given removal rate. (2) For the retained blocks, we apply a specially designed group parameter sharing strategy, where blocks within the same group share identical weights, significantly compressing the number of parameters and slightly reducing latency overhead. (3) After sharing these Blocks, we "cure" the mismatch caused by sparsity with a minor amount of fine-tuning and introduce a tail-layer distillation strategy to improve the performance. Experiments demonstrate that \textbf{FoldGPT} outperforms previous state-of-the-art(SOTA) methods in efficient model compression, demonstrating the feasibility of achieving model lightweighting through straightforward block removal and parameter sharing.
\end{abstract}

\section{Introduction}


\begin{figure*}[!htb]
    \centering 
    \includegraphics[width=\textwidth]{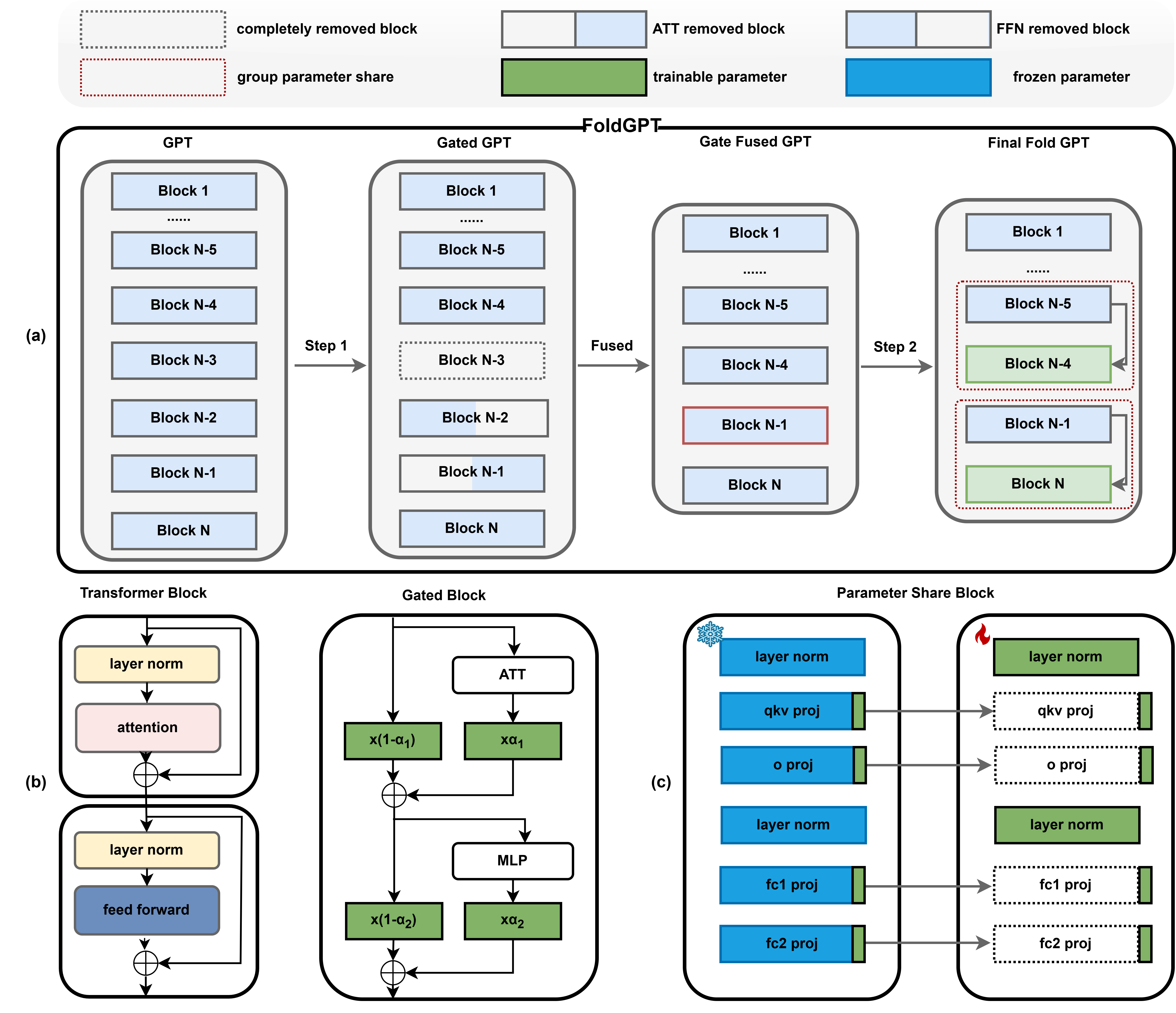} 
    \caption{An overview of our \textbf{FoldGPT}. (a)Two-step volume compression strategy, including gated block removal and grouped parameter sharing. (b) Block structure with learnable gating parameters. (c) Grouped parameter sharing structure. The first block in the group is called the parent block, and the remaining blocks share the weight parameters of the parent block, which are called child blocks.} 
    \label{fig:overview} 
\end{figure*}

\label{Section1}
Recently, LLMs have garnered tremendous attention in the field of artificial intelligence, which can be attributed to the success of models such as ChatGPT\cite{brown2020language}. 
Following scaling laws\cite{kaplan2020scaling}, leading models(like OPT\cite{zhang2022opt}, BLOOM\cite{le2023bloom}, and LLaMA\cite{touvron2023llama}) tend to increase in size to improve performance.
However, this substantial growth in model scale poses significant challenges for cloud deployment. MobileLLM\cite{liu2024mobilellm} points out that if 5\% of human individuals' time will use LLM services and call GPT-4 at a speed of 50 Tokens/s, then 100 million H100 GPUs would be needed to meet the computing requirements. The ensuing energy consumption and carbon dioxide emissions would present staggering environmental challenges. 

To tackle these challenges, researchers have begun deploying LLMs on smart-phones and mobile devices.
This approach leverages edge computing power to reduce the cost of cloud inference and protect user privacy. 
However, mobile devices are constrained by limited computing power, main memory (DRAM) capacity, and power supply. 
For example, the DRAM capacity of mainstream flagship phones ranges from 6GB to 12GB, and APPs can only utilize 10\% to 20\% of it\cite{liu2024mobilellm}. Therefore, the model volume has become a core issue hindering the deployment of LLMs on mobile devices. 
Researchers have proposed a variety of techniques to alleviate the computational burden and model size of LLMs while preserving their performance, including model pruning\cite{wang2019structured}\cite{xia2022structured}\cite{kurtic2022optimal}\cite{ma2023llm}, quantization\cite{frantar2022gptq}, distillation\cite{sun2019patient}\cite{wang2020minilm}, etc. 
Quantization methods reduce the memory access requirements of the model by decreasing the weight and activation bit width, allowing adaptation to new hardware computing units to achieve acceleration and storage compression benefits\cite{lin2024qserve}. Pruning algorithms obtain lighter and more hardware-friendly models by directly removing unimportant parameters from the model. 
Previous LLMs pruning research focused on analyzing the redundancy in the width of the model\cite{zhu2023survey}. Parameter reduction was achieved by removing head in the Self-Attention module and channels in the FFN module, which incurred high retraining costs\cite{ma2023llm}. 

In this paper, we analyze the similarity of activation values across each block in LLMs and find that many blocks contribute minimally, indicating substantial redundancy in the model's depth.
This finding was also demonstrated by ShortGPT\cite{men2024shortgpt}, a work at the same time. 
To address this deep redundancy, we propose a learning-based volume compression strategy for pre-trained models, called \textbf{FoldGPT}, which consists of two main steps: block removal and block parameter sharing. As shown in Figure~ \ref{fig:overview}(a), in the first step, the importance of each block is measured by a learnable gating parameter. 
With minimal fine-tuning, we can quickly get the importance ranking of the blocks, and remove redundant layers based on the given removal rate. In the second step, for reserved blocks, we apply a specially designed group parameter sharing strategy, which means that multiple blocks will share the same set of weight parameters. This approach significantly reduces the overall parameter amount and improve cache hits rate\cite{liu2024mobilellm}. Finally, we introduce additional learnable parameters to each shared block, which do not affect the computational efficiency, and then through distillation fine-tuning, 
we quickly restore the performance loss caused by parameter sharing. To the best of our knowledge, \textbf{FoldGP} is the first framework that integrates block removal and parameter sharing for ultimate LLM volume compression. The main contributions of our paper are summarized as follows:
\begin{itemize}
\item We analyze the similarity of block outputs in LLMs, with parameter sizes ranging from hundreds of megabytes to billions, proving that LLM possess extensive redundancy in their depth. This finding indicates that model pruning can be achieved by directly removing redundant blocks.
\item We propose a learnable block importance evaluation criterion and an adaptive polarization fine-tuning strategy. 
Unlike the BI\cite{men2024shortgpt} metric used by ShortGPT, our approach models the interactions between blocks, resulting in superior performance.
\item We propose a group parameter sharing strategy for pre-trained models. The blocks within a group share the same set of weights, supplemented by a few 
additional learnable parameters, and then a specially designed tail-layer distillation strategy to improve the performance.
\item we conduct extensive experiments on multi-scale models: LLaMA-2-7B, Gemma-2B, TinyLLaMA-1.1B. The experimental results demonstrate that even with the removal of 36\% of the parameters, the pruned model maintains 96.25\% of the performance of the original model, surpassing the current SOTA pruning algorithm. 
\end{itemize}


\begin{figure*}[!htb]
  \centering
  \begin{subfigure}[b]{0.3\textwidth}
    \includegraphics[width=\textwidth]{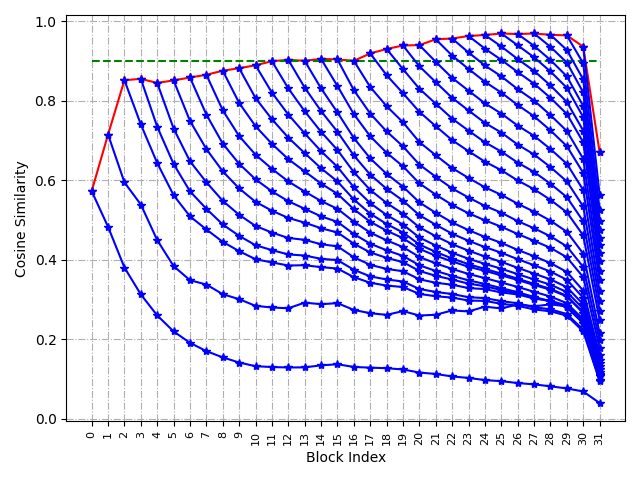}
    \caption{LLaMA-2-7B}
    \label{fig:llamav2-7b}
  \end{subfigure}
  \hspace{0.5cm}
  \begin{subfigure}[b]{0.3\textwidth}
    \includegraphics[width=\textwidth]{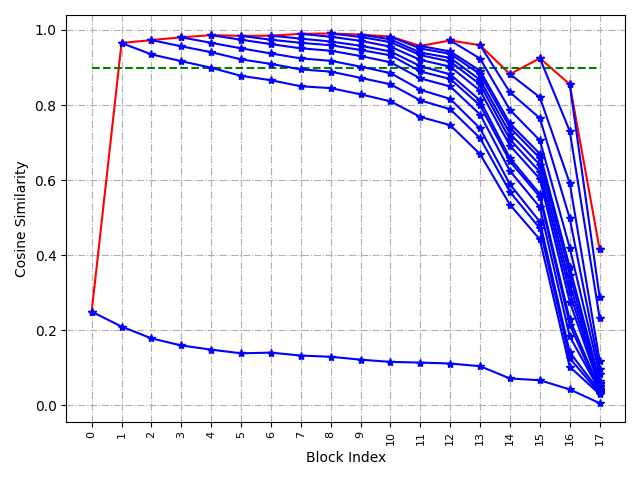}
    \caption{Gemma-2B}
    \label{fig:gemma-2b}
  \end{subfigure}
  \hspace{0.5cm}
  \begin{subfigure}[b]{0.3\textwidth}
    \includegraphics[width=\textwidth]{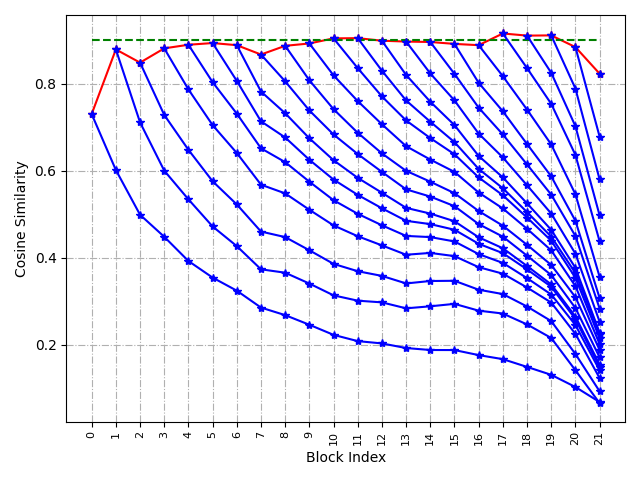}
    \caption{TinyLLaMA-1.1B}
    \label{fig:tinyllama}
  \end{subfigure}
  \caption{Block redundancy analysis of models with sizes from 1B to 7B. The \textcolor{red}{red} line represents the cosine similarity of the input and output of the current Block, while the \textcolor{blue}{blue} line represents the cosine similarity of the output of the current Block and the input of the starting point.}
  \label{fig:cos}
\end{figure*}

\section{Related works}
The substantial number of parameters in high-performance models necessitates considerable storage and memory, rendering these models unsuitable for devices with limited computing resources. Consequently, model compression has become a promising solution to this issue. In this section, we provide a brief overview of related works.

\paragraph{Quantization} 
Quantization can be classified into two main categories: Quantization-Aware Training (QAT)\cite{liu2023llm,ma2024era} and Post-Training Quantization (PTQ) \cite{dettmers2208llm,frantar2022gptq,xiao2023smoothquant,lin2023awq}. However, in the field of LLMs, due to the significant training costs, both the academia and industry mainly focused on PTQ methods. Among them, \cite{dettmers2208llm} proposed LLM.int8(). They first used vector-wise quantization for most weight matrix. To the emergent outliers, they incorporated a novel mixed-precision decomposition scheme, isolating the outlier feature dimensions into a 16-bit matrix multiplication. Along with LLM.int8(), GPTQ \cite{frantar2022gptq} proposed a more accurate data-aware approach via an approximate large-scale solver for minimizing layer-wise l2 errors. Further, SmoothQuant\cite{xiao2023smoothquant} not only quantized both the weights and activations, but also offline migrating the quantization difficulty from activations to weights with a mathematically equivalent transformation. Numerous experiments have demonstrated that PTQ methods have become the leading  techniques for compressing and deploying large-scale LLMs efficiently.

\paragraph{Pruning And Parameter Sharing}
Pruning techniques aim to identify and remove redundant or less significant parameters from models, resulting in a sparser weight matrix. These methods utilize heuristics, such as magnitude-based pruning, or more sophisticated approaches like learning-based pruning, to determine which weights to eliminate. Previous research has mainly analyzed model redundancy in terms of width. For example, SparseGPT \cite{frantar2023sparsegpt} incorporates the Hessian inverse for pruning and subsequent residual weight updates, while Wanda \cite{sun2023simple} achieves a sparse LLM model by using a criterion based on the product of the absolute values of weights and their activations to preserve outliers. \cite{ma2023llm} obtains a lightweight model by removing redundant heads in self-attention and redundant channels in FFN, followed by low-cost fine-tuning to restore accuracy. In contrast, our \textbf{FoldGPT} and ShortGPT \cite{men2024shortgpt} exploit model depth redundancy to obtain lightweight models. Specifically, \textbf{FoldGPT} uses learnable gating parameters to model the coupling between blocks and employs a parameter-sharing strategy to compress effective parameters. The basic idea of parameter sharing is to use the same set of parameters for multiple parts of an LLM. In existing research \cite{ullrich2017soft, lin2023understanding, su2024beyond, liu2024understanding}, parameter-sharing strategies are primarily used in model infrastructure design and pre-training to enhance computational efficiency and reduce overfitting risk, especially with limited data. For instance, ALBERT \cite{liu2024understanding} uses a cross-layer parameter-sharing strategy to effectively reduce the number of model parameters, achieving better training results than the baseline with the same parameter number \cite{su2024beyond}. Furthermore, \cite{liu2024mobilellm} confirms that models employing parameter sharing generally achieve better performance during pre-training compared to those without sharing. Unlike these approaches, we apply parameter sharing to compress the trained model. By integrating a specialized sharing module and a distillation fine-tuning strategy, we achieve state-of-the-art performance.

\paragraph{Distillation}
Knowledge Distillation (KD) \cite{hinton2015distilling} is widely used to transfer knowledge from a large model (teacher) to a smaller one (student) for improved efficiency. In the context of LLMs, KD preserves the rich semantic and contextual understanding of these models. Previous works use soft target probabilities or intermediate representations from the teacher model to guide the task-agnostic student model training. For example, DistilBERT \cite{sanh2019distilbert} halves the transformer's layers in the teacher network, initializing the student by selecting one layer out of every two from the teacher. This ensures the student retains the teacher's architecture. TinyBERT \cite{jiao2019tinybert} and MobileBERT \cite{sun2019mobilebert} transfer fine-grained knowledge, including hidden states and self-attention distributions, using linear layers or bottleneck modules for alignment. In contrast, MiniLM \cite{wang2020minilm} simplifies the process by distilling knowledge solely from the self-attention module of the last Transformer block, alleviating the challenge of layer mapping. Block removal and group parameter sharing based on the pre-trained model introduce additional performance degradation. Inspired by MiniLM, we employ q-k-v distillation of the last block during fine-tuning. Experimental results show that this strategy effectively enhances the performance of the compressed model.

\section{Methodology} \label{Section3}
\subsection{Redundancy Analysis} \label{Section31}
Mainstream LLMs are usually stacked by repeated decoding blocks, known as Transformers\cite{lagler2013gpt2}. As shown in Figure~\ref{fig:overview}(b), a standard decoding block contains an attention module(ATT) and a feed-forward module(FFN). For an L-layer transformer-based LLM, the input of i-th layer is denoted by $X_{i} \in\mathbb{R}^{ b \times s \times d} $, where $b$, $s$ and $d$ respectively represent the batch size, number of tokens, and hidden dimensions. The input of layer $i+1$ can be expressed as follows:

\begin{small}
\begin{equation}\label{equation1}
\begin{aligned}
    X^{i}_{ATT}&= X^{i} + Attention(LN(X^{i})) \\
    X^{i+1}&= X^{i}_{ATT} + FFN(LN(X^{i}_{ATT}))
\end{aligned}
\end{equation}
\end{small}
Since stacked decoding blocks typically have the same structural parameters, this means that $X^{i}$ and $X^{i+1}$ share the same dimensions, prompting us to investigate the role each block plays in the information flow path. As shown in Figure~\ref{fig:cos}, 
we selected three models: LLaMA-2\cite{roumeliotis2023llama}, Gemma\cite{team2024gemma}, and TinyLLaMA\cite{zhang2024tinyllama}, with parameter sizes of 1.1B, 2B, and 7B respectively, to analyze the cosine similarity of different activation values within the model. The red lines in the three sub-figures reflect the importance of a individual block without considering inter-block coupling, while the blue line models the importance of groups of consecutive blocks. Our analysis yielded following findings: (1) The cosine similarity of most intermediate blocks in the LLMs is greater than 0.9, indicating their minimal role in the information transmission path. (2) In LLaMA-2-7B and Gemma-2B, the cosine similarity of activations across multiple consecutive blocks remains above 0.8, suggesting the potential for grouped block removal. Group block redundancy decreases with the model size decreases. For example, under the criterion of cosine similarity greater than 0.8, TinyLLaMA can only remove two consecutive blocks. Based on these findings, we propose \textbf{FoldGPT}, a novel model compression strategy, that combines block removal and grouped block parameter sharing. The former targets a small number of particularly redundant blocks, and the latter targets relatively lower redundancy block group. In contrast, ShortGPT\cite{men2024shortgpt} only focuses on the redundancy of individual block and cannot perform grouped block parameter removal due to significant performance degradation.

\begin{figure}[t]
  \small
  \centering
  \begin{subfigure}[b]{0.4\textwidth}
    \includegraphics[width=\columnwidth]{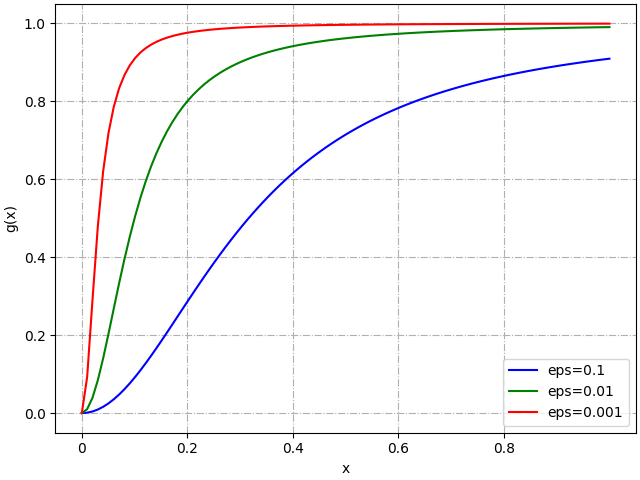}
    \caption{Forward under different $eps$}
    \label{fig:gx}
  \end{subfigure}
  \hspace{0.5cm}
  \begin{subfigure}[b]{0.4\textwidth}
    \includegraphics[width=\columnwidth]{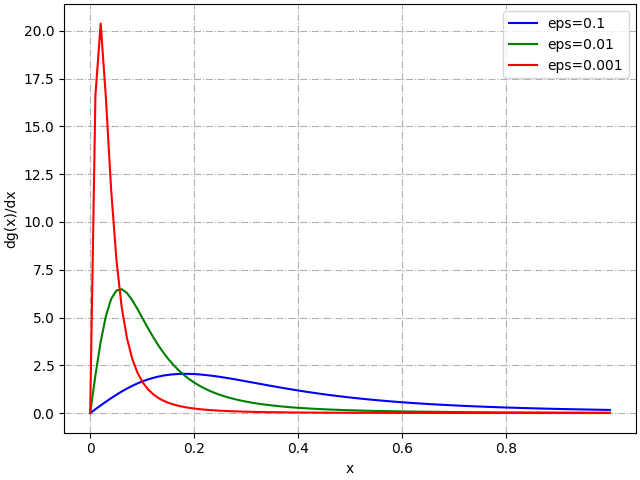}
    \caption{Backward under different $eps$}
    \label{fig:dgx}
  \end{subfigure}
  \caption{Polarization characteristics on different $eps$.}
  \label{fig:gdp}
\end{figure}

\begin{table}
    \centering
    \caption{Perplexity of WikiText2 and PTB on LLaMA-2-7B with different pruning rates.}
    \label{tab::gate_ppl}  
    \begin{tabular}{cccc}
        \toprule  
        Ratio& Method & WikiText2$\downarrow$ & PTB$\downarrow$\\ \hline
        0 & - & 5.5012 & 20.642\\ \hline
        15\% & BI& 10.4837& 32.980\\
        & \textbf{FoldGPT} & \textbf{7.430} & \textbf{24.537} \\
        \hline
        27\% & BI& 35.962& 85.896\\
        & \textbf{FoldGPT} & \textbf{18.500} & \textbf{55.939} \\
        \bottomrule 
    \end{tabular}
    \vspace{-0.5 em}
\end{table}

\subsection{Gated block removal}
\label{Section32}

In the first step of \textbf{FoldGPT}, we directly remove certain blocks according to a specified sparsity rate, which means that we need to accurately evaluate the importance of the blocks and sort them. ShortGPT\cite{men2024shortgpt} uses Block Influence(BI), which is essentially cosine similarity, to measure the importance of each block. However, this metrics has a significant flaw: it does not account for the coupling between blocks, leading to sub-optimal solutions when removing multiple blocks. To address this issue, we propose a gated block removal strategy, which introduces a learnable gating parameter for each minimal component. This approach enables a more accurate ranking of the importance of inter-block coupling through joint optimization. As shown in Figure~\ref{fig:overview}(b), we reformulate the calculation process of a block as follows:

\begin{small}
\begin{equation}\label{equation2}
\begin{aligned}
    X^{i}_{ATT}= X^{i}*(1-g(\alpha^{i}_{1})) + ATT(X^{i})*g(\alpha^{i}_{1}) \\ X^{i+1}= X^{i}_{ATT}*(1-g(\alpha^{i}_{2})) +  FFN(X^{i}_{ATT})*g(\alpha^{i}_{2})
\end{aligned}
\end{equation}
\end{small}

\noindent where $\alpha^{i}_{1}$ and $\alpha^{i}_{2}$ respectively represent the gating parameter of the ATT module and FFN module in the i-th decoding Block, and $g(.)$ denotes the gating activation function. The introduction of an additional gating function aims to encourage the gate values to update smoothly towards polarization during training, such that some values converge to exactly zero while others remain significantly different from zero. This process allows us to obtain an accurate ranking of block importance based on the gate values after training. Inspired by \cite{guo2021gdp}, we employ the smoothed L0 formulation from principled optimization methods to achieve this goal:

\begin{small}
\begin{equation}\label{equation3}
\begin{aligned}
    g_{eps}(x) =\frac{x^2}{x^2+eps} \quad \\ \quad \phi(g_{eps}(x)) =\frac{2\times x \times eps}{x^2+eps} \\ 
\end{aligned}
\end{equation}
\end{small}

\noindent where $g_{eps}(x)$ and $\phi(g_{eps}(x))$ represent the forward and backward calculation of the gate function, respectively, where $eps$ is a hyper-parameter that balances training stability and polarization. As shown in Figure~\ref{fig:gdp}, we can observe that when $eps$ is sufficiently small, $g_{eps}(x)$ becomes polarized, and $\phi(g_{eps}(x))$ is zero only at specific points. This characteristic suggests that we can use $g_{eps}(x)$ as a gating mechanism, integrating it  into the neural network we ami to compress. In our study, $eps$ is initialized to 0.1 and gradually decreases during the training process. 
This approach ensures both early training stability and the polarization of gating parameters by the end of training. 
To further enhance the polarization of gating parameters, we also introduce resource constraints into the optimization function. In our paper, in order to simplify the problem, we directly use FLOPs(Floating Point Operations) as the measure of resource consumption for each block and consider retaining or removing the ATT and FFN modules simultaneously. For an L-layer LLM, the final objective function can be formulated as below:

\begin{small}
\begin{equation}
	\begin{aligned}
	\underset {\alpha} { \operatorname {min} } \Bar{\mathcal L}(W;\alpha) = \mathcal L(W; g(a)) + \lambda\sum_{0}^{L-1} g(a^{i})S^{i}
\end{aligned}
\end{equation}
\end{small}
\noindent where $\mathcal L$ is the original loss function, and  $\alpha=\{\alpha^{0},...,\alpha^{L-1}\}$ represents the final importance score of each block, and $\lambda$ is a balance factor. Since there are few parameters to optimize, the optimization process is efficient. Obtain the importance ranking based on the gating parameters at step 1K and directly compare it with BI-based sequence.
As shown in Table~\ref{tab::gate_ppl}, we directly apply the removal sequence obtained through the gated learning strategy without any fine-tuning and compare it with BI metrics of ShortGPT\cite{xiao2023smoothquant}. 
Our strategy significantly outperforms BI. At a removal rate of 15\%, the perplexity(ppl) on the WikiText2 dataset is reduced by 29\%, and the perplexity on the Penn Treebank(PTB) dataset is reduced by 25\%. When the pruning rate further increased to 27\%, the advantages further expand to 48.5\% and 34.88\%. 

\subsection{Grouped parameter sharing}
\label{Section33}
In the second step of \textbf{FoldGPT}, we implement group parameter sharing for the remaining blocks. This approach further compress the amount of model parameters. Although it does not reduce the computational load, it can theoretically enhance performance on devices with limited storage resources due to the improved cache hit rate \cite{liu2024mobilellm}. Figure~\ref{fig:overview}(c) illustrates the block parameter sharing details for a group size is 2. Specifically, each child block contains 4 fully connected layers whose weight parameters reused from the corresponding layers of the parent block. In order to improve the accuracy of fitting the child block weights using the parent block weights, we introduce additional learnable scaling coefficients for the weights in each child block. Assume that the i-th weight in a child block is $W^{i}_{child}\in \mathbb R^{d_{out}\times d_{in}}$, where $d_{out}$ and $d_{in}$ represent output and input channels respectively. Then, the forward computation can now be expressed as follows:
\begin{equation}
	\begin{aligned}
    W^{i}_{child} = W^{i}_{parent}*S^{i}_{child} \\
    f^{i}_{child} = (W^{i}_{parent}*S^{i}_{child}) * X + B
\end{aligned}
\end{equation}
\noindent where $W^{i}_{parent} \in \mathbb R^{d_{out}\times d_{in}}$, $S^{i}_{child} \in \mathbb R^{d_{out}} $ and $f^{i}_{child}$ represent the corresponding the reused weights, the additional scaling parameters introduced by the child layer, and the forward process of the child layer, respectively. When performing the forward calculation of the child layer, the scaling parameter does not introduce additional calculation time because it can be seamlessly integrated into the tail processing. Additionally, we apply two techniques to improve the model performance: \textbf{layernorm re-adaptation} and  \textbf{parent weight fine-tuning}. This involves fully opening the parameters of the layer normalization (LN) layer and partially opening the weights of the parent dense layer during subsequent fine-tuning training. To demonstrate the effectiveness of  these technique, we conducted ablation experiments on Gemma-2B and TinyLLaMA-1.1B. As shown in Table~\ref{tab::gate_ln_father}, enabling LN training can effectively improve the generation quality. The perplexity on the WikiText2 drop by 2.13 and 9.61, respectively, while the perplexity on PTB drop 18.3 and 24.13, respectively.
\begin{table}
    \centering
    \small
    \caption{Comparison of perplexity when the layernorm layer is frozen and unfrozen.}
    \label{tab::gate_ln_father}  
    \begin{tabular}{cccc}
        \toprule
        Model & Ratio & WikiText2 & PTB \\ \hline
        & 0\% & 8.93 & 28.97 \\
        Gemma & 26.34\% wo ln & 30.52 & 199.15 \\
        & 26.34\% w ln & \textbf{28.39} & \textbf{180.85} \\ \hline
        & 0\% & 7.93 & 19.39 \\
        TinyLLaMA & 36\% wo ln & 80.43 & 223.14 \\
        & 36\% w ln & \textbf{70.82} & \textbf{199.01} \\ \bottomrule
    \end{tabular}
\end{table}

\subsection{Distillation fine-tuning}
\label{Section34}
After completing the two stages of parameter compression, we need to restore the accuracy through fine-tuning. In order to expedite the model recovery process and enhance efficiency under limited data conditions, we employ the low-rank approximation(LoRA)\cite{hu2021lora} for post-training the pruned model. Notably, during the fine-tuning, only the dense layer in the parent block and the layer normalization in the child block are activated. LoRA is only applied to the weight of the dense layer. The forward calculation of the sub-layer under LoRA can be expressed as follows:

\begin{small}
\begin{equation}
	\begin{aligned}
    f^{i}_{child} = ((W^{i}_{parent}+\triangle W^{i}_{parent})*S^{i}_{child}) * X + B
\end{aligned}
\end{equation}
\end{small}

\noindent $\triangle W^{i}_{parent}$ is the updated value of the parent layer weight, which can be decomposed as  $\triangle W^{i}_{parent} = P*Q $, where $P\in \mathbb R^{d_{out} \times d_{-}}$ and  $Q\in \mathbb R^{d_{-} \times d_{in}}$. Since $d_{-}$ is much smaller than $d_{in}$ and $d_{out}$, this decomposition reduces training complexity and the need for large-scale training data.
In order to further speed up fine-tuning and improve model performance, we introduce a knowledge distillation strategy inspired by \cite{wang2020minilm}. Due to structural misalignment between the original model and the compressed model after block removal and folding, we apply the distillation loss only to the last layer. Specifically, the distillation loss consists of two components: self-attention and self-attention value relation, represented by $\mathcal L(\triangle W)_{AT} $ and $\mathcal L(\triangle W)_{VR}$, respectively. The final loss can be expressed as:

\begin{small}
\begin{equation}
	\begin{aligned}
    \hat{\mathcal L }= \mathcal L + \lambda (\mathcal L(\triangle W)_{AT} + \mathcal L(\triangle W)_{VR})
\end{aligned}
\end{equation}
\end{small}

\begin{table*}[!htb] \small
\centering
\caption{Zero-shot Performance on LLaMA-2-7B. * indicates that the model has been fine-tuned with the same configuration. Detailed hyper-parameters are described in section\ref{Section412}}
 \label{tab::main_results}  
\begin{tabular}{ccccccccc}
\toprule
\multirow{2}{*}{Model} & \multirow{2}{*}{Method} & \multirow{2}{*}{Ratio} & \multicolumn{4}{c}{Benchmarks} & \multirow{2}{*}{Avg.} & \multirow{2}{*}{Per.} \\ 
 &  &  & HeSw & PIQA & BoolQ & MMLU &  &  \\ \hline
\multirow{10}{*}{LLaMA-2-7B} & Dense & 0.00\% & 71.26 & 77.91 & 71.62 & 45.39 & 66.55 & 100\% \\ 
 & LLMPruner & 27.0\% & \textbf{56.46} & \textbf{71.22} & 55.20 & 23.33 & 51.55 & 77.46\% \\
 & SliceGPT & 26.4\% & 50.27 & 66.21 & 38.32 & 28.92 & 45.93 & 69.01\% \\
 & LaCo & 27.1\% & 55.69 & 69.80 & 64.07 & 26.45 & 54.00 & 81.14\% \\
 & ShortGPT & 27.1\% & 53.02 & 66.43 & 74.71 & 43.96 & 59.53 & 89.45\%\\
 & \textbf{FoldGPT} & 27.1\% & 53.22 & 67.00 & \textbf{73.21} & \textbf{45.14} & \textbf{59.64} &   \textbf{89.61\%} \\ \cline{2-9} 
 & ShortGPT* & 27.1\% & 57.7 & 67.70 & 73.20 & 43.70 & 60.57 & 91.01\% \\
  & \textbf{FoldGPT*1}  & 27\%+9\%  & 59.67  & 68.85 & 75.20 & 43.88 & 
 61.90 & 93.01\%\\
  & \textbf{FoldGPT*2}  & 21\%+15\%  & 61.25  & 73.22 & \textbf{75.20} & 44.04 &
 63.41 & 95.28\%\\
 & \textbf{FoldGPT*3}  & 15\%+21\%  & \textbf{63.10}  & \textbf{74.34} & 74.30 & \textbf{44.50}& 
 \textbf{64.06} & \textbf{96.25\%}\\ \bottomrule
\end{tabular}
\end{table*}

\section{Experiments}
\subsection{Experimental Setup}
\label{Section41}
\noindent \textbf{Model}
\label{Section411}
To comprehensively evaluate the performance of our algorithm, we selected LLaMA-2-7B \cite{touvron2023llama}, Gemma-2B \cite{team2024gemma}, and TinyLLaMA-1.1B \cite{zhang2024tinyllama} as our base models. 
Gemma-2B is a lightweight, state-of-the-art open model series developed by Google, based on the same research and technology as the Gemini model. TinyLLaMA-1.1B is pre-trained on approximately 3 trillion tokens, leverages the LLaMA-2 architecture and tokenizer. Its small size makes it suitable for applications with limited computing and memory resources. 
Given that smaller models typically exhibit lower redundancy, comparisons involving these models are particularly meaningful for assessing the effectiveness of our algorithm.

\noindent \textbf{Training}
\label{Section412}
During the gating parameter removal phase, we initialize $eps$ to 0.1 and decay it by 0.97 decay rate every 120 iterations. During training, all model parameters are frozen, with only gate control parameters are released.
In the recovery phase, we utilize the cleaned version of Alpaca dataset\cite{alpaca}, which comprises approximately 50k samples. For LoRA training, we set the LoRA rank $d_{-}$ to 8 and initialize the distillation loss coefficient $\lambda$ to 1e-5. The learning rate is set to 1e-5 with 100 warming steps. The training batch size is selected from \{32, 64\} and the AdamW optimizer\cite{zhuang2022understanding} is employed in our experiments. 
Additionally, we found that the optimal training duration is 2 epochs, as training for more epochs even has a negative impact on the model performance. Our experiments were conducted on a single NVIDIA A100-GPU with 80GB memory, taking approximately 3 hours.

\begin{table*}[!htb] \small
\centering
\caption{Zero-shot Performance on Gemma-2B. * indicates that the model has been fine-tuned with the same configuration. Detailed hyper-parameters are described in section\ref{Section412}}
 \label{tab::gemma}  
\begin{tabular}{cccccccccc}
\toprule
\multirow{2}{*}{Model} & \multirow{2}{*}{Method} & \multirow{2}{*}{Ratio} & \multicolumn{5}{c}{Benchmarks} & \multirow{2}{*}{Avg.} & \multirow{2}{*}{Per.} \\ 
 &  &  & WinGran & SCIQ & RACE & PIQA & BOOLQ &  &  \\ \hline
\multirow{6}{*}{Gemma-2B} & Dense & 0.00\% & 65.50 & 90.21 & 36.26 & 76.93 & 69.41 & 66.44 & 100\% \\ 
 & ShortGPT*1 & 13.17\% & 52.22 & 74.9 & 25.55 & 65.58 & 52.61 & 54.17 & 81.53\% \\
 & ShortGPT*2 & 17.56\% & 51.01 & 69.7 & 24.49 & 63.27 & 50.70 & 51.83 & 78.02\%\\
 & ShortGPT*3 & 21.95\% & 48.51 & 68.8 & 25.26 & 61.61 & 49.04 & 50.64 & 76.23\% \\ \cline{2-10} 
  & \textbf{FoldGPT*1}  & 8.78\%+8.78\%  & 57.69  & 88.1 & 33.49 & 71.43 & 62.87 & 
 \textbf{62.72} & \textbf{94.39\%}\\
  & \textbf{FoldGPT*2}  & 17.56\%+4.39\%  & 54.77  & 78.2 & 28.42 & 63.76 & 60.58 &
 \textbf{57.15} & \textbf{86.00\%}\\
 & \textbf{FoldGPT*3}  & 17.56\%+8.78\%  & 54.69  & 78.7 & 27.94 & 63.54 & 62.04 & 
 \textbf{57.38} & \textbf{86.37\%}\\ \bottomrule
\end{tabular}
\end{table*}

\noindent \textbf{Evaluated Tasks}
Following the previous work, we evaluate the model based on two indicators: perplexity and zero-shot common sense reasoning score. Perplexity is calculated based on the WikiText-2 \cite{merity2016pointer} and Penn Treebank(PTB)\cite{marcus1993building} datasets, with a truncation length of 2048 tokens per sample. 
For evaluating the performance on zero-shot common sense reasoning tasks, we select serveral popular tasks including BoolQ \cite{clark2019boolq}, PIQA \cite{Bisk2020}, HellaSwag \cite{zellers2019hellaswag}, SCIQ \cite{welbl2017crowdsourcing} , ARC \cite{clark2018think}, WinoGrande \cite{ai2:winogrande} and MMLU\cite{hendrycks2020measuring}. We adhere to the GPTQ settings\cite{frantar2022gptq} for our language generation experiments and utilize the lm-eval-harness framework\cite{gao2021framework} for executing all zero-shot tasks.

\subsection{Main Results}
\label{Section42}

To validate the advantages of our algorithm, we compared \textbf{FoldGPT} with state-of-the-art structured pruning algorithms, including LLMPruner\cite{ma2023llm}, SliceGPT\cite{ashkboos2024slicegpt}, LaCo\cite{yang2024laco} and ShortGPT\cite{men2024shortgpt}. LLMPruner and SliceGPT primarily exploit redundancy in network width by compressing embedding dimensions, while Laco and ShortGPT focus on removing redundancy in network depth. To ensure fairness in the absence of fine-tuning in the comparative study, we first applied block removal based on a gating mechanism to LLaMA-2-7, achieving a sparsity of 27\%. As shown in Table~\ref{tab::main_results}, compared to the width-based structured pruning algorithms SliceGPT and LLMPruner, \textbf{FoldGPT} showed significant improvements of 13.71 and 8.9, respectively. This indicates that depth-based structured pruning outperforms traditional width-based pruning in LLMs,supporting our view that depth redundancy is higher than width redundancy. In comparison with LaCo and ShortGPT, \textbf{FoldGPT's} gating-based block removal strategy significantly outperformed LaCo's layer-wise removal strategy, demonstrating the substantial advantage of our gating mechanism in block importance analysis. To further explore \textbf{FoldGPT's} extreme compression rate under group parameter sharing and ensure a fair comparison, we applied the same fine-tuning strategy to both ShortGPT and \textbf{FoldGPT}. The results showed that \textbf{FoldGPT} achieved a performance improvement of 1.33 over ShortGPT with 9\% group parameter sharing. Furthermore, maintaining a sparsity rate of 36\%, \textbf{FoldGPT} achieved a significant performance improvement of 5.24\% over ShortGPT by flexibly allocating the proportions of block removal and group parameter sharing, even with 9\% additional sparsity compared to ShortGPT.

Finally, to investigate the performance advantage of \textbf{FoldGPT} in edge deployment, we conducted a detailed comparison with the current leading structured pruning algorithm, ShortGPT, using the Gemma-2B model.  As shown in Table~\ref{tab::gemma}, we applied a combination of 17.56\% block removal and 4.39\% parameter folding. Compared to ShortGPT, which utilized only 17.56\% block removal, \textbf{FoldGPT} improved benchmark accuracy by 5.31\% and increased the compression rate by 4.39\%. Furthermore, in the comparison between \textbf{FoldGPT*2} and \textbf{FoldGPT*3}, \textbf{FoldGPT*3} achieved results comparable to \textbf{FoldGPT*2} by further increasing the block folding rate, while also demonstrating an 8.35\% performance improvement and an 8.73\% increase in sparsity over ShortGPT. These findings validate our approach in edge LLM models and further confirm the advantages of \textbf{FoldGPT} over other structured pruning algorithms.

\section{Conclusion}
\label{Section5}
In this paper, based on the similarity of each block outputs, we propose an efficient LLMs compression method, \textbf{FoldGPT}, to address the challenges of deploying LLMs on mobile devices. 
\textbf{FoldGPT} first proposes a learnable block importance evaluation criterion and an adaptive polarization fine-tuning strategy, which can remove the relatively unimportant blocks more reasonable.For the remained blocks, \textbf{FoldGPT} then proposes a group parameter sharing strategy with a small number of additional learnable parameters, which can further compresses the LLMs footprint.For alleviating the performance loss resulted from the above two strategies, \textbf{FoldGPT} introduces 
a specially designed tail-layer distillation strategy to improve the performance. Extensive experimental results demonstrate that \textbf{FoldGPT} could outperform the SOTA pruning algorithms with respect to the same pruning ratio.

\section*{Limitations}
\label{Section6}
The \textbf{FoldGPT} can reduce footprint usage and improve inference speed for more efficient deployment of LLMs. However, there are still some limitations. 
Firstly, \textbf{FoldGPT} is proposed for the LLMs where the basic blocks are stacked repeatedly. As a result, this strategy cannot be applied the for the LLMs, i.e., OpenELM\cite{mehta2024openelm} where the structural parameter configuration of each block is different.
Secondly, due to the large granularity of parameter folding, there exists some loss to model performance, so we need to cooperate with fine-tuning to restore accuracy.
Thirdly, in this paper, we mainly focus on the implementation of algorithms because of time constraints. In the future, we will carry on implementing the corresponding inference engine and gain actual performance acceleration benefits.

\section*{Ethics Statement}
This paper presents solutions to the challenges of pruning Large Language Models (LLMs) to facilitate their widespread adoption and application. Currently, ethical concerns related to LLMs, such as hidden biases encoded in the models, are receiving increased attention. Our investigation indicates that our proposed method does not amplify these biases or contravene any ethical standards.

\bibliography{custom}

\appendix



\end{document}